\documentclass[conference]{IEEEtran}

\usepackage{graphicx}
\usepackage{multirow}
\usepackage{rotating}
\usepackage{hyperref}

\ifCLASSINFOpdf
\else
\fi

\usepackage{booktabs} 

\begin{document}
%
\title{A Comparison of Lexicon-Based and ML-Based Sentiment Analysis: Are There Outlier Words?}

\author{\IEEEauthorblockN{Siddhant Jaydeep Mahajani$^{1}$, Shashank Srivastava$^{1}$ and Alan F. Smeaton$^{1,2}$}
\IEEEauthorblockA{$^{1}$School of Computing and $^{2}$Insight Centre for Data Analytics\\
Dublin City University, Glasnevin, Dublin 9, Ireland\\
email: alan.smeaton@dcu.ie}}

\maketitle

\begin{abstract}
Lexicon-based approaches to sentiment analysis of text are based on each word or lexical entry having a pre-defined weight indicating its sentiment polarity. These are usually manually assigned but the accuracy of these when compared against machine leaning based approaches to computing sentiment, are not known.  It may be that there are  lexical entries whose sentiment values cause a lexicon-based approach to give results which are very different to a machine learning approach.
In this paper we compute sentiment for more than 150,000 English language texts drawn from 4 domains using the Hedonometer, a lexicon-based technique and  Azure, a contemporary machine-learning based approach which is part of the Azure Cognitive Services family of APIs which is easy to use. We  model  differences in sentiment scores between  approaches for documents in each domain using a  regression and analyse the independent variables (Hedonometer lexical entries) as indicators of each word's importance and contribution to the  score differences. Our findings are that the importance of a word depends on the domain and  there are no standout lexical entries which systematically cause differences in sentiment scores.
\end{abstract}


%
\IEEEpeerreviewmaketitle

\section{Introduction}

Sentiment analysis is an established form of text analysis which measures to what extent the sentiment behind a piece of text is positive, negative or neutral.
The most popular implementation  combines machine learning and natural language processing (NLP) though one of the downsides is domain-dependence
where classifiers need to be attuned to different text domains. An alternative approach is lexicon-based using a dictionary of words with pre-defined sentiment ratings. This has the advantage of domain-independence but brings a disadvantage and a commonsense assumption that the semantics of a word should  depend on itself and also  its use and context. The use of word context can give significant improvements on a wide range of NLP tasks including sentiment analysis.

Here we are interested in how domain-independent are lexicon-based sentiment analysis tools, how do the ``baked-in" word-level sentiment weights contribute to differences when compared to machine learning approaches, and how transferable are they across domains?  We take a popular lexicon-based sentiment analysis tool, the Hedonometer \cite{frank_happiness_2013}, and compare its output  against that from a popular machine learning based tool, Microsoft Azure's Text Analytics technology \cite{carvalho2020off} on  collections of text from four  domains.  We set the Azure sentiment analysis as a  standard to aim at and we use a  regression to model the differences in sentiment analysis  from the two approaches across each of the four domains. We then examine significance values for the variables from the regression thus revealing what are the lexical entries, i.e.,  words which have the greatest and  least impact on differentiating between Hedonometer and Azure sentiment scores. This highlights  Hedonometer words  whose sentiment weights may need to be updated and those which should be left untouched if we wanted Hedonometer sentiment  to match  Azure sentiment, for each domain.  
The insights this provides will help us understand the strengths and limitations of lexicon-based approaches to calculating sentiment scores.

\section{Background and Related Work}

\subsection{Sentiment Analysis}

Sentiment analysis, or opinion mining, is the automatic analysis of text in order to determine the attitude or judgement that the text prompts in a typical reader \cite{neethu2013sentiment}. It
has widespread use in social media monitoring, brand monitoring and reputation management, product analysis and customer reviews, and in market research \cite{wankhade2022survey}.
The most popular implementation  combines machine learning and one of its sub-fields, NLP, on manually annotated training data to achieve systems which are robust and scalable \cite{ain2017sentiment}.
One of the downsides of such approaches is domain-dependence
where
``linguistic and content peculiarities require a domain-specific sentiment source'' \cite{DENECKE201517}.

\subsection{Lexicon-Based Sentiment Analysis}

An alternative approach to using machine learning is word or lexicon-based where the 
polarity of a text is determined by searching for words or phrases which have pre-determined weights as indicators of sentiment,  then combining the word weights  in some way. 
Lingmotif is a lexicon-based, linguistically-motivated, sentiment analysis  tool \cite{moreno-ortiz-2017-lingmotif} which   performs analysis on  input text based on the identification of sentiment-laden words and phrases from Lingmotif’s rich core lexicons. It also employs context rules to account for sentiment shifters.
SentText is another  tool for lexicon-based sentiment analysis \cite{schmidt2021senttext} which performs sentiment analysis with predefined sentiment lexicons or self-adjusted lexicons.
Finally, Syuzhet is a lexicon-based tool for  sentiment analysis of literary texts that draws upon the Syuzhet, Bing, Afinn, and NRC lexicons \cite{kim2022sentiment} containing 10,748, 6,789, 2,477 and	13,901 words respectively.

Despite  shortcomings, lexicon-based sentiment  methods are widely used.
The methods have often been criticised for their accuracy but  recent work \cite{ohman-2021-validity} has shown that lexicon-based methods can work well  where neither qualitative analysis such as manually assigned ground truth labels, nor a machine learning-based approach, is possible.

\subsection{Hedonometer}

The Hedonometer is a lexicon-based sentiment analysis tool which measures average ``happiness'' or sentiment   using a  lexicon of 10,222 common words in the English language, each of which has a context-free estimation of its ``happiness'' score. These scores were calculated using a language assessment Mechanical Turk  where users  were asked to rate each word on a nine-point integer scale according to how it made them feel \cite{frank_happiness_2013}.
Examples of the averaged  scores for some  words from the Hedonometer are shown in Table~\ref{tab:scores}.

\begin{table}[ht]
\centering 
    \caption{Sample from the 10,222 words and their   scores in the range  1 (sad) to 9 (happy) from \cite{dodds2011temporal}}
  \label{tab:scores}
  \begin{tabular}{lc|lc|lc}
\toprule
Word & Score & Word & Score &  Word & Score \\
\midrule
laughter & 8.50 &food & 7.44 & reunion & 6.96  \\
the & 4.98 & of & 4.94 & vanity & 4.30  \\
hate & 2.34 & funeral & 2.10  & terrorist & 1.30 \\
\bottomrule
\end{tabular}
\end{table}

\noindent 

Since its introduction, the Hedonometer has demonstrated stability and reliability along with a remarkable quality of tunability \cite{dodds2011temporal}. The algorithm to compute sentiment for a document initially  extracts the frequencies of occurrence and then the average  sentiment from a given text which is subsequently normalised for document length.

Many lexicon-based sentiment analysis tools such as Hedonometer  have  limitations as they fail to  consider word context. For instance, the phrase \textit{not happy} would receive a  positive sentiment score, but the phrase \textit{not unhappy} would receive a  negative one. Hence  we can say that Hedonometer, like most lexicon-based approaches to sentiment analysis, should not be very reliable in calculating sentiment scores for short texts but if  errors in sentiment are not connected then this may not be an issue when it is used on a large quantity of text  \cite{gibbons_2019}.

\section{Data Collections}

For the analysis of Hedonometer vs. Azure sentiment analysis we used annotated English language data sets from four  domains: Finance, News, Social Media, and IMDb customer reviews. The premise  is that  data  from different domains helps us to identify the most commonly used words in that domain, i.e., domain-specific set of words as well as the words that are common irrespective of what their domain is.

{\bf Finance:}
Finance is a domain where sentiment is important as it can influence stock market trends. We used data from \cite{Malo2014GoodDO} who used it to identify semantic orientations in economic texts. It consisted of 
c.5,000 phrases/sentences sampled from financial news texts and company press releases,  tagged as positive, negative or neutral. 

{\bf News:}
The  news data set  was  
used by \cite{dangi2023efficient} to perform sentiment analysis on  news that are displayed everywhere. It consists of almost 50,000 news articles for an 8 month period from November 2015 to July 2016 on four different topics: economy, Microsoft, Obama and Palestine.

{\bf Social Media:}
The social media data was  a collection of 40,000 tweets and used by \cite{barbieri2020tweeteval} to perform sentiment analysis on  Tweets  posted by  users with the specific task of emoji prediction. 

{\bf IMDb Reviews:}
The IMDb customer reviews consisted of 50,000 reviews posted on   IMDb,   an International Movie Database platform. The data set was used by \cite{maas-etal-2011-learning} to perform sentiment analysis.

The number of Hedonometer lexical entries appearing across  texts in each domain is shown in Table~\ref{tab:hedowords} showing that texts from the news domain have much fewer Hedonometer words. 
However the reader is reminded that our objective is to see if it is possible to identify lexical entries in the Hedonometer which cause it to differ from a machine learning based approach and not to determine the ultimate adjustments to Hedonometer weights that should be enacted.

\begin{table}[htb]
    \centering
        \caption{Charactistics of Hedonometer words from the lexicon of 10,222,  appearing in  datasets from each domain}
    \begin{tabular}{lccccc}
    \toprule
     &Finance & News & Soc. media & IMDB  & All domains \\
     \midrule
     \# Hedonometer  &966    & 274   & 1,886   & 2,673  & 3,810   \\
     words &&&&&\\
     \bottomrule
     \\
    \end{tabular}
    \begin{tabular}{cccc}
    \multicolumn{4}{c}{Numbers of words appearing in any of}\\
    1 domain  & 2 domains  & 3 domains  &  all domains\\
    \midrule 
    2,396  & 941   & 371 & 102 \\ 
    \bottomrule 
    \end{tabular}
    \label{tab:hedowords}
\end{table}

\begin{figure*}[htb]
    \centering
    \includegraphics[width=0.49\textwidth]{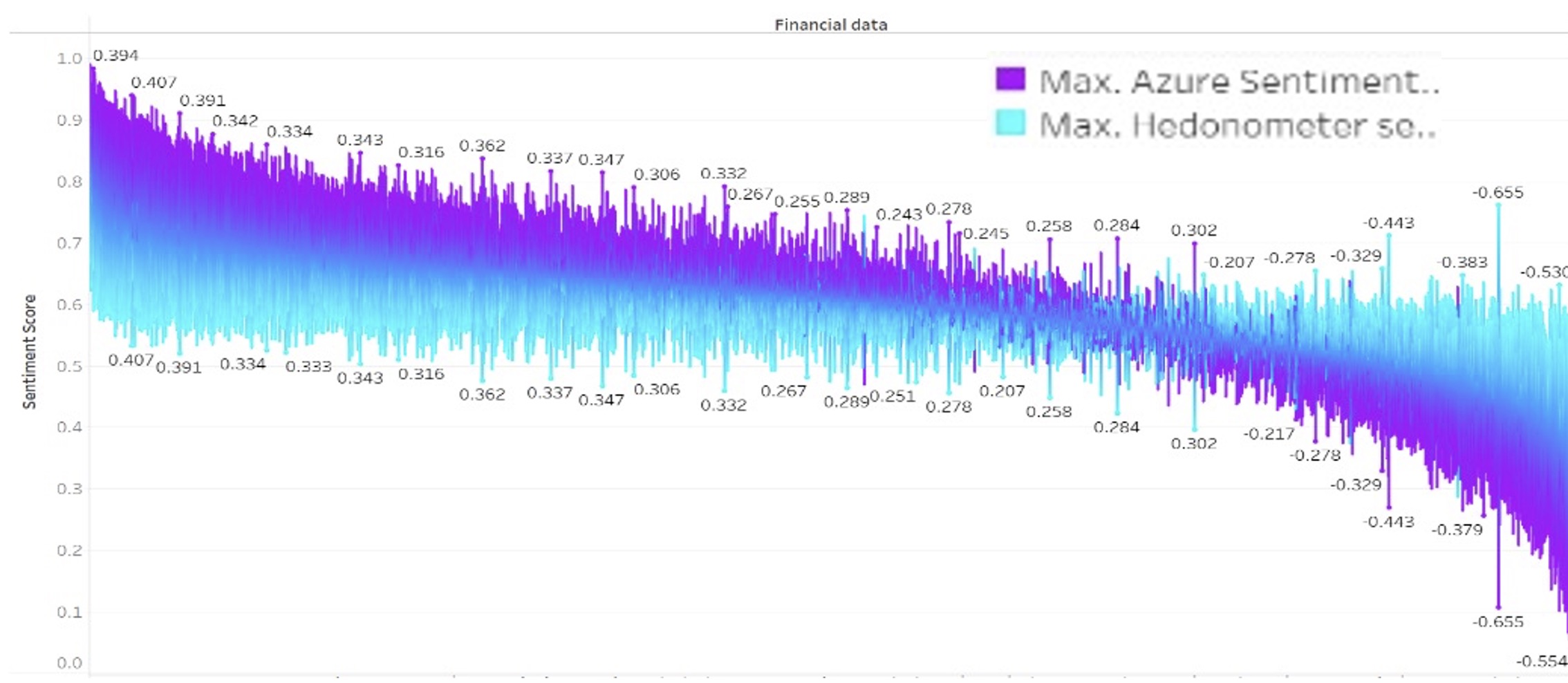}
    \includegraphics[width=0.49\textwidth]{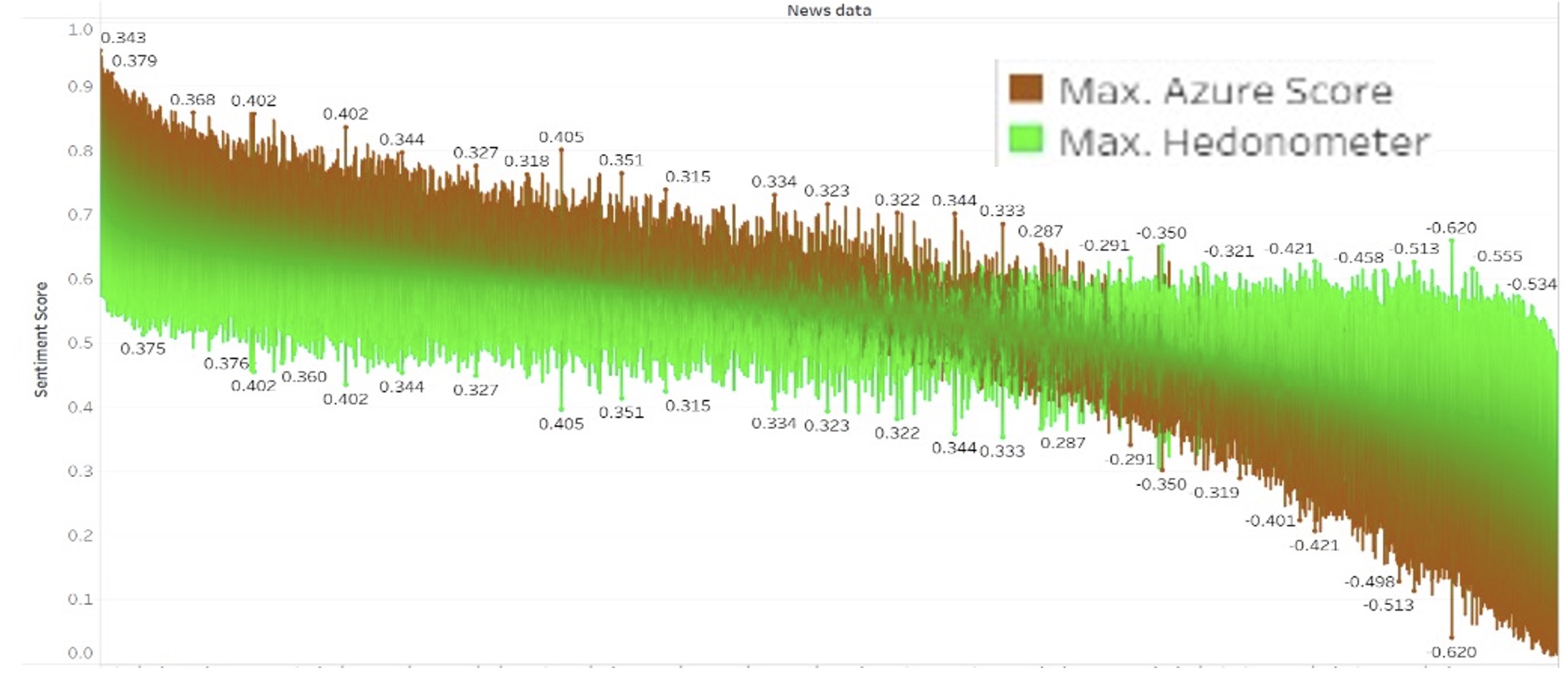}
    \includegraphics[width=0.49\textwidth]{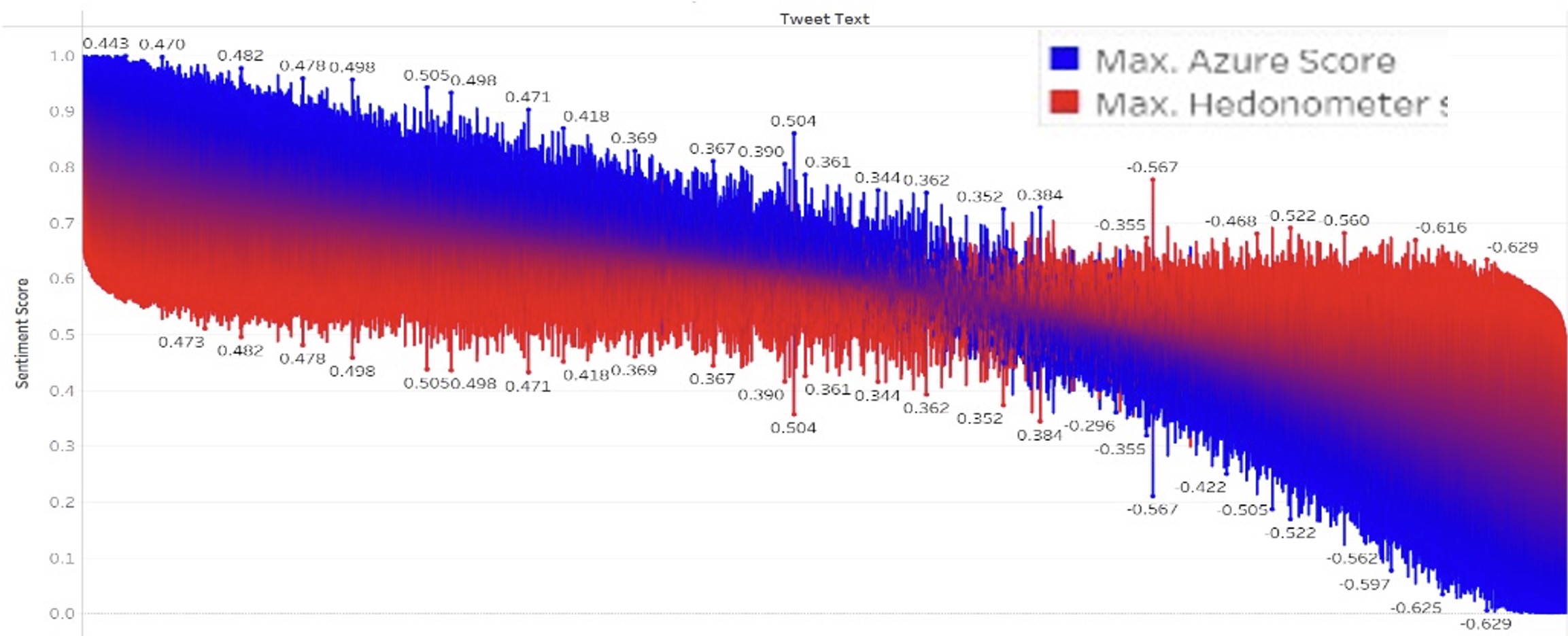}
    \includegraphics[width=0.49\textwidth]{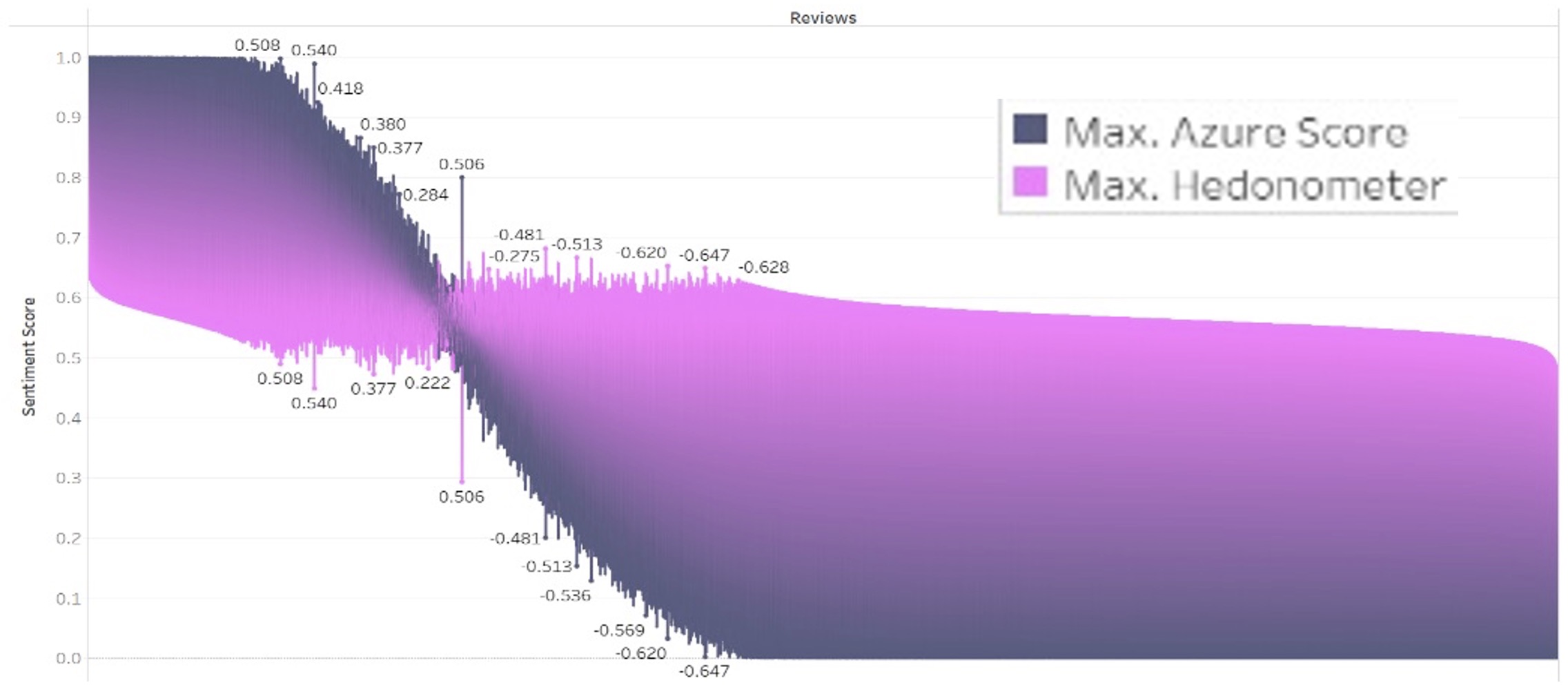}
    \caption{Differences between Hedonometer  and Azure sentiment  scores   for   finance, news, social media and IMDB review domains, respectively.}
    \label{fig:distributions}
\end{figure*}

\subsubsection*{Calculating Sentiment Scores}

we used Microsoft Azure's Text Analytics technology \cite{carvalho2020off}, an established machine-learning based sentiment analysis tool, against which to match Hedonometer scores. This scores texts in the interval  0 (negative) to 1 (positive) for sentiment. 
We computed the sentiment score for each document in each domain using Hedonometer and Azure and Figure~\ref{fig:distributions} shows, for each  domain, the differences in scores. For  finance and  news  there is a relatively flat part of the graph in the middle indicating approximate agreement between Hedonometer and Azure, with small numbers of documents at each end where there are larger differences in sentiment. For  social media documents there are greater differences between Hedonometer and Azure  ratings while for the IMDB reviews there are extreme differences  with few documents having agreed or even close scores (the crossover part of the graph). This is caused by a relatively small range of  values from the Hedonometer (shown in pink) while the Azure ratings have a much greater range of  values. 

\section{Results and Analysis}
 
We set the Azure sentiment scores as a target  and  use linear regression to model the differences in results between the Hedonometer  and Microsoft Azure's Text Analytics technology across texts from each of the four domains. We then examine  significance values for  variables from the regression  revealing which words have  greatest and which have  least impact on differentiating between Hedonometer and Azure. 
The top half of Table~\ref{tab:p-vals}  lists   words from each domain and from a combination across all domains, with the smallest  p-values indicating  words whose contributions are different between Hedonometer and  Azure.  
These are words whose sentiment weights would need to be changed to make the Hedonometer more like Azure.  The bottom half of Table~\ref{tab:p-vals}  lists   words
with the largest p-values indicating  words with the same interpretations in Hedonometer as in Azure. We limit our analysis  to words which appear in Hedonometer's lexicon and in at least three of the four text domains to see if there are words consistently outliers across domains.  

\begin{table*}[!ht]
\centering 
    \caption{Hedonometer lexicon words with smallest and largest  p-values when modelling    Hedonometer vs. Azure outputs.}
  \label{tab:p-vals}
  \begin{tabular}{ccccc|c}
\toprule
&\multicolumn{4}{c|}{Individual domains} & {Combined domains} \\
\cline{1-5}
&Finance & News &  Social media & IMDB reviews &  ranked by p-value \\
\midrule
\multirow{10}{*}{\begin{turn}{90}Smallest p-values\end{turn}}
&jason & great &  great &  listen  & understand  \\
&sign  & mom  & wish &  video   & walk \\
&strong  & water  & may &  dream   & bad \\
&point  & mail  &  feel   & rigid  & meant \\
&profit  & video  & wait &  hang   & end \\
&fan  & press  & want &  ad   & goodnight \\
&tough  & strength  & still &  known   & space \\
&owl  & bomb  & mom &  avoid   & main \\
&matt  & earn  & kind &  violent   & faith \\
&rate  & flag  & found &  laura   & sister \\
\midrule 
\multirow{10}{*}{\begin{turn}{90}Largest p-values\end{turn}}
&mon & taken &  louis &  new  & high  \\
&greg  & small  & water &  use   & sport \\
&notion  & consider  & town &  editor   & sign \\
&co  & india  &  ipad   & charter  & blank \\
&blank  & reach  & bout &  paid   & upset \\
&pilot  & chose  & demon &  written   & worst \\
&shall  & worst  & Monday &  snake   & new \\
&take  & jordan  & top &  role   & good \\
&bear  & ok  & round &  ten   & best \\
&report  & upset  & friend &  sweden   & opinion \\
\bottomrule
\end{tabular}
\end{table*}

We  measured the correlation between  rankings of (some) Hedonometer words by their  ``happiness'' scores vs. the p-values from differences between Hedonometer and Azure. Table~\ref{tab:spearmans} shows those correlations with words whose p-values equal to 0 removed.  
This shows a negative correlation for Finance and News and a positive but not strong correlation  for the others . What this means is that the importance of a word towards the differences between the two sentiment analysis approaches has only a small correlation with the Hedonometer ranking of that word.

\begin{table}[htb]
    \centering
    \caption{Correlation between Hedonometer  score ranking and ranking by p-value }
    \begin{tabular}{ccc}
    \toprule
    Domain & Number of Hedonometer & Spearman correlation \\
    &words used  &\\
    \midrule
Finance & 808 & -0.1126 \\
News & 259 & -0.1206 \\
Social media & 967 & 0.2949 \\
IMDB reviews & 1,369 & 0.3300 \\
\bottomrule
    \end{tabular}
    \label{tab:spearmans}
\end{table}

\section{Discussion and Conclusions}

Though the number of Hedonometer words in each of the four domains may   limit our analysis, restricting it to entries from just a portion of the possible lexicon, this does not detract from the process of trying to identify lexical entries  which cause it's output to differ from the Azure machine learning approach.
Table~\ref{tab:p-vals} indicates that when mapping Hedonometer sentiment  to Azure sentiment, different words are more, and less, important for different domains.  This may be because the sets of Hedonometer words appearing in the texts from the four domains are  a fraction of the overall Hedonometer lexicon, 3,810 from 10,222 possible  entries. Even so, there are no major outlier words that stand out which is surprising, but informative and the only word with smallest or largest p-value that appears in more than one domain is ``mom".  Our Spearman correlation between Hedonometer ranking and p-values for modelling the differences, bears this out. The nature of the words in Table~\ref{tab:p-vals} do not appear to be particularly domain-dependent and from those words it would be difficult to match them to their domain.

Our future work may include targeting the impact of specific Hedonometer words for their impact on sentiment analysis by using texts containing those words rather than using words from particular domains as we have done here.  This would give us greater coverage than the 3,810 words we analysed here though the results may be the same.
There may also be an issue around our use of linear regression to determine p-values in the situation where the many independent variables in the model, the lexical entries, lead to multiple hypothesis testing. While
correction methods for multiple hypothesis testing exist \cite{menyhart2021multipletesting} which we could use in future, an alternative would be to substitute the regression model with a non-parametric machine learning model where feature importance scores could be used to evaluate the importance of individual lexical entries.

Considering that there is no such thing as universally agreed sentiment scores \cite{doi:10.1177/0165551517703514} and even inter-annotator agreement among humans is only about 80\%, adjustments to the weights of Hedonometer lexical entries would seem to make little or no difference to overall sentiment scores.  In the actual use of sentiment analysis tools, so long as they are used consistently and any comparisons are like-with-like and not across sentiment analysis tools or approaches then modifications to weights in lexicon-based approaches may not be worthwhile.

\vspace{10pt}

\noindent 
{\bf Acknowledgment}
This work was part-funded by Science Foundation Ireland under Grant Number SFI/12/RC/2289\_P2, co-funded by the European Regional Development Fund. Data  is  available at \url{https://doi.org/10.6084/m9.figshare.24539410}.



\bibliographystyle{IEEEtran}
\bibliography{AICS2023.bib}
%



\end{document}